\documentclass[conference]{IEEEtran}
\IEEEoverridecommandlockouts
\usepackage{cite}
\usepackage{amsmath,amssymb,amsfonts}
\usepackage{algorithmic}
\usepackage{graphicx}
\usepackage{textcomp}
\usepackage{xcolor}
\def\BibTeX{{\rm B\kern-.05em{\sc i\kern-.025em b}\kern-.08em
    T\kern-.1667em\lower.7ex\hbox{E}\kern-.125emX}}
\begin{document}

\title{Hybrid Trajectory Optimization of Simple Skateboarding Tricks through Contact\\
}

\author{\IEEEauthorblockN{Michael J. Burgess}
\IEEEauthorblockA{\textit{Department of Electrical Engineering and Computer Science (EECS)} \\
\textit{Massachusetts Institute of Technology (MIT)}\\
Cambridge, MA, USA \\
mjburgessjr@gmail.com}
}

\maketitle

\begin{abstract}
Trajectories are optimized for a two-dimensional simplified skateboarding system to allow it to perform a fundamental skateboarding trick called an "ollie". A methodology for generating trick trajectories by controlling the position of a point-mass relative to a board is presented and demonstrated over a range of peak jump heights. A hybrid dynamics approach is taken to perform this optimization, with contact constraints applied along a sequence of discrete timesteps based on the board's position throughout designated sections of the trick. These constraints introduce explicit and implicit discontinuities between chosen sections of the trick sequence. The approach has been shown to be successful for a set of realistic system parameters.
\end{abstract}


\section{Introduction}
An "ollie" is a fundamental skateboarding trick which requires a rider to shift their weight backwards on the board to kick it up into the air \cite{b1}. After jumping into the air, a rider shifts their weight forward to land the board front first. The rotation of the board is manipulated via the contact forces on the rider's feet. These forces change in magnitude based on the position of the rider's body \cite{b2}. The trick occurs in two necessary dimensions, going up and down within a plane orthogonal to the width of the board.

An ollie is used to jump into the air and over obstacles when riding a skateboard. The most critical part of an ollie is the kick-off \cite{b3}. This is when the rider shifts their weight fully backward such that the tail end of the board collides into the ground, reversing motion upwards and lifting the rider into the air. Kick-off is where the board-rider skateboarding system transitions from being in-contact with the ground to fully out of contact. Stronger kick-off allows for a greater peak height of the jump to be achieved \cite{b3}. 

A robotic skateboarding system could be developed to perform tricks such as an ollie, enabling it to navigate complex street environments. Being able to perform an ollie would allow a skateboarding system to jump obstacles autonomously. No work has been found attempting to achieve a similar goal. Finding optimal ollie trajectories could also be used to better inform how the trick is performed and how it could be better performed to maximize jump height. Previous studies have investigated the dynamics of ollie jumps and highlighted the critical nature of landing and kick-off in jumping mechanics \cite{b3}\cite{b4}. In the following paper, a methodology to find optimal ollie trajectories for a simplified underactuated robotic skateboarding system over a smooth theoretical terrain will be described.

\section{Methods}
\subsection{System}

A simplified two-dimensional system can be used to model an ollie trajectory. Previous studies have modeled skateboard dynamics using a point mass to represent the rider. Moving this mass relative to the board can be used to manipulate its orientation and subsequent motion \cite{b5}. To perform an ollie, a similar approach can be taken in two-dimensions. In this case, a point mass can be moved relative to the board within the two-dimensional plane. Changing the position of this mass is analogous to the rider shifting their weight back and forth across their feet. A description of this proposed of this system can be seen in Fig. 1, with parametrized lengths of skateboard geometry and angles.

\begin{figure}[htbp]
\centerline{\includegraphics{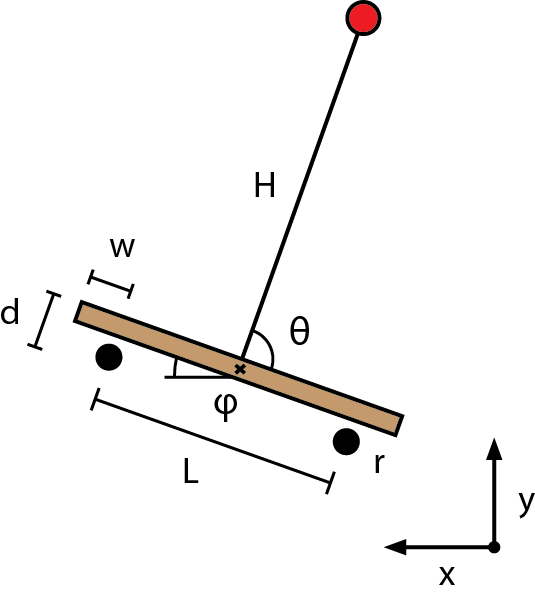}}
\caption{The proposed two-dimensional skateboard system with point mass rider $m_r$ and board mass as $m_b$. Variables in the figure describe geometry of the board, which can be assumed as a rigid body. Wheel radius is represented with $r$.}
\label{fig}
\end{figure}

The orientation and position of the board's center of mass are considered the state of the system $\bar{q}$. This state is manipulated via the control variable $\bar{u}$, which is equivalent to the angular position of the point mass relative to the board. The point mass is placed at a distance of $H$ from the board's center of the mass, this length should be made equal to the relative hip-height of the rider.
\begin{align}
\bar{q} &= \begin{bmatrix}
       x \\
       y \\
       \phi
     \end{bmatrix},\quad
\bar{u} = \begin{bmatrix}
       \theta
     \end{bmatrix}
\end{align} 
\\
The system is assumed to be actuated via an ideal stepper motor that perfectly controls the angle of the point mass rider relative to the board. The torque from this motor imparts a moment on the board which could cause it to spin, affecting $\phi$. When in full contact with the ground, the board experiences reaction forces upon each wheel, front and back. These reaction forces are labeled as $R_1$ and $R_2$ respectively. Both masses experience a gravitational force downwards. Free body diagrams for both the simplified rider and board are shown in Fig 2.

\begin{figure}[htbp]
\centerline{\includegraphics{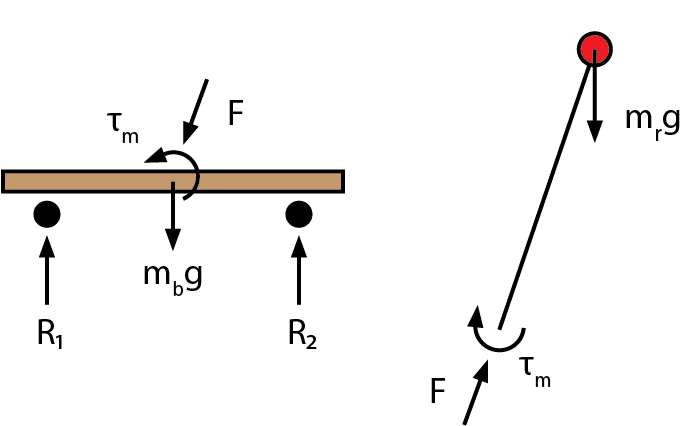}}
\caption{Free body diagrams on rider and board subsystems. Axial reaction force between subsystems shown as $F$. Reaction moment imparted from motor is displayed as $\tau_m$. Reaction forces are non-zero and positive when the board is in contact with the ground.}
\label{fig}
\end{figure}

System dynamics of state variables are described below. They have been found to be nonlinear and dependent on the sine and cosine of difference between angles. The rigid pole holding the point mass rider away from the board is considered massless and in tension. The mass of the rider should be about an order of magnitude higher than that of the board. Friction is neglected because the mass of the skateboard's wheels can be considered much less than that of the rider. Furthermore, the terrain which is rides along is considered to be smooth.\begin{multline}
(m_r + m_b)\ddot{x} = \\ 
= -m_{b}h((\ddot{\theta} - \ddot{\phi})\text{sin}(\theta - \phi) + (\dot{\theta} - \dot{\phi})^2\text{cos}(\theta - \phi))
\end{multline}
\begin{multline}
(m_r + m_b)\ddot{y} = R_1 + R_2 - (m_r + m_b)g + \\
+ m_{b}h((\dot{\theta} - \dot{\phi})^2\text{sin}(\theta - \phi) - (\ddot{\theta} - \ddot{\phi})\text{cos}(\theta - \phi))
\end{multline}
\begin{multline}
(\frac{m_b(L + 2w)^2}{12} + \frac{m_{r}h^2}{2})\ddot{\phi} = \\
= \frac{L}{2}(R_1 - R_2) + \frac{m_{r}h}{2}(g\text{cos}(\theta - \phi) + h\ddot{\theta})
\end{multline}

\subsection{Optimization Through Contact}

An ollie trajectory can be optimized to analyze the point mass rider should be moved in order to jump over a specified height. A direct transcription approach will be taken, with control input $\bar{u}$ and state $\bar{q}$ and their respective derivatives as decision variables. Hybrid trajectory optimization has been shown as a strong method for planning in systems that come in and out of contact with the ground \cite{b6}. Reaction forces will also be optimized as decision variables. They should be contrained to zero when the board is out of contact with the ground, and positive when in contact. These force constraints introduce implicit discontinuities in dynamics because they are discontinuous along section breaks. Optimization can be performed over $T$ discrete timesteps, deciding the duration between these discrete steps with an additional decision variable $h$. This style of approach has been outlined and proven feasible in analogous walking systems \cite{b7}. Total discrete timesteps $T$ are divided into 5 sections based on modes of contact as shown in Fig. 3. Over every section, linear constraints are added to ensure that the height $y$ of the board and orientation $\phi$ correspond to the expected behavior within that section.

\begin{figure}[htbp]
\centerline{
    \includegraphics[width=250pt]{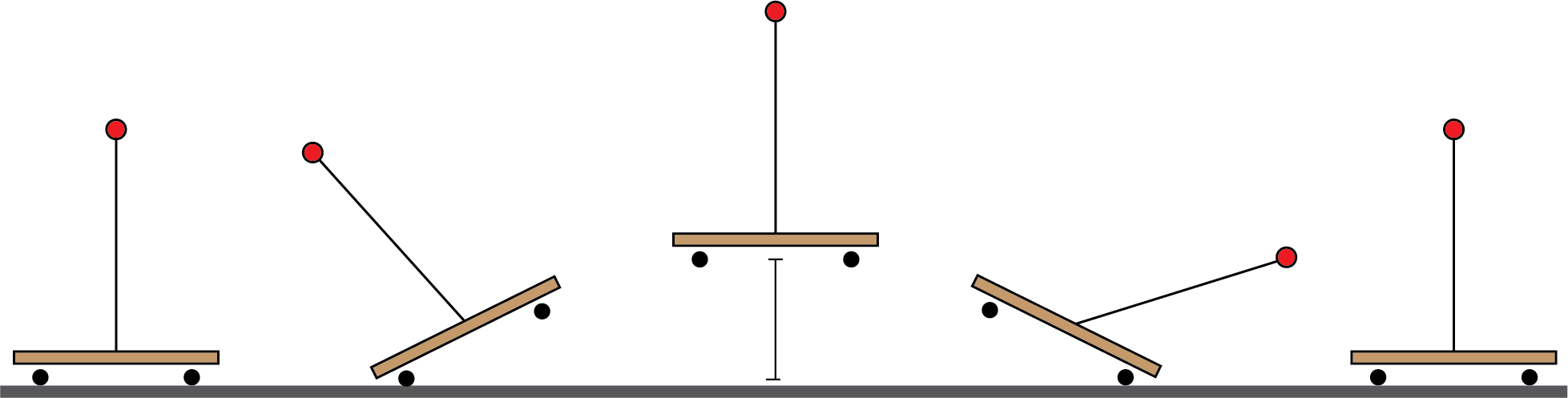}
}
\caption{Contact modes broken into 5 sections along the duration of the jump, starting from the right and moving to the left. The skateboard system begins on the ground, reach a peak height in the air, and descends front-first.}
\label{fig}
\end{figure}

At the start of the trick, the board will be flat upon the ground such that it's $y$ coordinate can be constrained to $d$ and orientation $\phi$ stuck at $0$. $\theta_0$ is constrained as $\pi/2$ to start the system in equilibrium. Here, both reaction forces $R_1$, $R_2$ should be restricted to positive values and start balanced. As the trick begins, weigh should be shifted towards the back wheel until $R_1$ approaches zero. Here, in the second section of timesteps, only the back wheel is in contact and the board begins to tilt back. $R_1$ remains zero in this section, since the front wheel is out of contact. The back wheel is constrained to remain on the ground and the board tilts upward along this axle like an angular joint, while still sliding along the ground as if there's a prismatic joint in the $x$ direction. This is held using the trigonometric nonlinear constraint outlined in Eq. 5.
\begin{align}
y_i = r + (d-r)\text{cos}(\phi_i) + \frac{L}{2}\text{sin}(\phi_i)
\end{align} 

When the board tilts upwards until the tail hits the ground, then it kicks off into the air. This kick-off occurs when $\phi$ has increased such that nonlinear Eq. 6 is satisfied. Before kick-off, $\phi$ should be lower than this value, forcing the tail to be above the ground.
\begin{align}
r + (d-r)\text{cos}(\phi_{i_{kickoff}}) = w\text{sin}(\phi_{i_{kickoff}})
\end{align}

Kick-off signifies the transition from section 2 to fully airborne section 3, where both reaction forces are zero. Kick-off can be modeled as an inelastic collision into the ground, causing an instantaneous discontinuity in state where angular velocity of the board reverses. Kinetic losses are assumed with a coefficient of restitution, $e < 1$. A guard function (Eq. 7) is used to model kick-off and transition between contact sections 2 and 3.
\begin{align}
\dot{\phi}_{i_{kickoff} + 1} = -e\dot{\phi}_{i_{kickoff}}
\end{align}

Height of jump is enforced by constraining jump height of the midpoint $y_{T/2}$ to be equal to a provided value. The board lands mirrored to how it came upwards, landing front first. In section 4, $R_1$ is constrained to be positive and $R_2$ to be zero. Reaction forces act as guard functions between sections, going from positive to zero based on whether or not the respective wheel is in contact with the ground. In this manner, they introduce implicit discontinuities in system dynamics. When the front wheel lands at the start of section 4, it is restricted to remain on the ground using a similar nonlinear constraint as defined in Eq. 6. The board can then act as an angular joint with monotonically decreasing $\phi$ and free $x$. When the back wheel hits the ground, $R_2$ is allowed to be positive and the point mass is shifted until the system is stable and has returned to its initial state with $u_T = \pi/2$.

With contact constraints enforced over each section of the ollie sequence and dynamics enforced at every timestep, trajectory optimization can be solved. Input values are limited to $0 \le u_i \le \pi$ such that the point mass rider cannot move below the board. The magnitude of control acceleration $\ddot{u}$ is also restricted to remain below a constant value to enforce realistic trajectories. Furthermore, distance between timesteps $h_i$ is constrained to be within a reasonable range. This generates a solvable optimization problem that can be used to find all decision variables.
\begin{multline}
    \quad\quad\quad\; \text{find}\quad q, u, R, h \\
\text{subject to}\quad\quad\quad \ddot{q_i} = f(q_i, \dot{q_i}, u_i, \dot{u_i}, \ddot{u_i})\\
\forall\; i \in [0, T] \quad\quad\; 0 \leq u_i \leq \pi\quad\quad\quad\quad\;\;\; \\
\quad\quad |\ddot{u}| \leq \ddot{u}_{max} \\
\quad\quad\quad\quad\quad\; h_{min} \leq h_i \leq h_{max} \\
\quad\quad\quad\quad\quad\quad\quad y_{T/2} = \text{jump\_height}\quad \\
\quad\quad\quad\quad\quad\quad\quad\; \dot{\phi}_{kickoff}^+ = -e\dot{\phi}_{kickoff}^-\\
\text{Contact constraints}\quad\quad\quad\;\;\;
\end{multline}

The optimization problem was coded in Python using Pydrake package. Constraints were programmed as described above the problem was optimized using the iterative IPOPT solver. This solver has been shown to perform well for large-scale nonlinear optimization problems \cite{b8}. Thus, it was chosen to find an optimal trajectory considering the system's nonlinear dynamics and multiple nonlinear contact constraints described above.

A URDF file representing the system was constructed based on board and rider geometry. State and control position and velocities over discrete timesteps were plugged into this file using Meshcat via Pydrake to visualize generated trajectories. This system was used purely for visualization. Configured simulation software at a resting state is shown in Fig 4.\begin{figure}[htbp]
\centerline{\includegraphics[width=150pt]{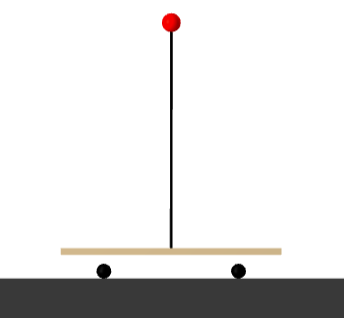}}
\caption{Screen capture of visualization software at a rest state with $y= d$, $\phi = 0$ and $\theta = \pi/2$. Geometric system parameters in this visualization are equal to those defined in Table 1.}
\label{fig}
\end{figure}

\section{Results \& Discussion}

Results were generated using the above described methodology for realistic system parameters. Rider mass was chosen to represent a 65 kilogram rider. Board parameters were chosen to represent a standard 32 inch long skateboard. These chosen parameters are defined in Table 1.
\begin{table}[htbp]
\caption{System Parameters}
\begin{center}
\begin{tabular}{ |c|c|c|c| } 
\hline
\;\;\;Variable\;\;\; & \;\;\;Value\;\;\; \\
\hline
$m_r$ & 65 kg \\ 
\hline
$m_b$ & 5.0 kg \\ 
\hline
$L$ & 0.50 m \\ 
\hline
$w$ & 0.16 m \\ 
\hline
$H$ & 0.85 m \\ 
\hline
$d$ & 0.10 m \\  
\hline
$r$ & 0.0275 m \\
\hline
$e$ & 0.80 \\
\hline
$T$ & 500 \\
\hline
\end{tabular}
\label{tab1}
\end{center}
\end{table}

Optimal trajectories were generated over a range of feasible jump heights using the above described methodology. State value $y$ is plotted for each of these jump heights over time in Fig 5. Control variable $u = \theta$ is plotted for these same trajectories in Fig 6.

\begin{figure}[htbp]
\centerline{\includegraphics[width=240pt]{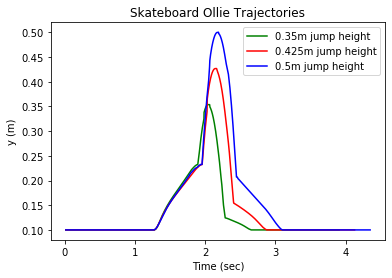}}
\caption{Optimal board height trajectories for a small range of jump heights plotted over time. Peaks achieved at the midpoint as constrained.}
\label{fig}
\end{figure}

As first obvious when viewing plotted trajectories of board height $y$, all peak jump heights are achieved as constrained. Jump height of 0.5m is the highest shown as trajectories with a higher designated height could not be solved using the described approach. This implies given system and contact constraints that there is some maximum jump height achievable. Future work could be done to try and tweak system parameters in order to increase this maximum height. Furthermore, it is evident that board lift-off during contact section 2 is nearly consistent for all tested jump heights. That is, board height raises in a very similar manner before kick-off. However, trajectories begin to deviate when kick-off occurs. This indicates that kick-off is a crucial point in the jumping trajectory as previously shown \cite{b3}. After kick-off, board center of mass raises rapidly until the board turns over and falls at a similar speed. Although peak heights are reached, the board only remains high in the air for a short period of time. Further work could be done to ensure the board stays in the air for a longer period of time, possibly by adding constraints on time or limiting elevation speed during airborne contact section.

\begin{figure}[htbp]
\centerline{\includegraphics[width=240pt]{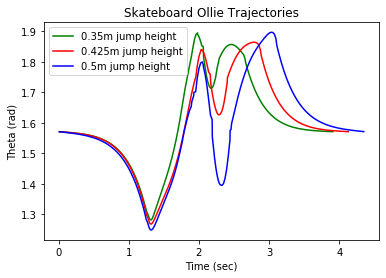}}
\caption{Optimal control input trajectories for a small range of jump heights plotted over time. Values start and end at stable position $u = \pi/2$ radians.}
\label{fig}
\end{figure}

As seen in plotted trajectories of board height $y$, control values remain consistent along the beginning stages of the trick until kick-off. At the point of kick-off, steeper control slopes are observed in trajectories that reach a higher peak height. This is likely due to the nature of the kick-off, where a higher control velocity leads to a higher angular velocity of the board, allowing it to kick-off at a higher speed. Visualized simulations re-affirm this observation. Most notably, during the flight phase where both wheels are out of contact with the ground, trajectories that reach a higher peak jump height demonstrate larger control recoil along the section of the trick sequence when the board begins to land. This implies that landing from a higher height requires larger control accelerations to re-balance the board, which makes sense because gravitational acceleration causes a larger ground impact velocity when falling from a higher point. The force of this impact may be difficult for the system to stabilize along for high jump heights. This observation may also reveal why trajectories for jump heights above 0.5m could not be found via the described optimization approach. Magnitude of control accelerations have been limited below a constant value of $\ddot{u}_{max} = 25$. Yet, these results imply that higher jump heights require larger control accelerations to properly land. Increasing this maximum value should allow for higher max jump heights. In future work, a better method of constraining inputs could be employed using a torque-speed curve of a feasible motor.

\section{Conclusion}

The designed simplified skateboarding system has been proven to be feasible to generate trajectories to perform an ollie using defined system parameters and the range of tested jump heights. Despite abstracting the rider to a controllable point mass, the skateboard was still able to perform tricks. Contact constraints were applied based on sections of the trick sequence over a number of discrete timestep, outlining a hybrid dynamics approach where implicit discontinuities in these constraints are enforced across section breaks. An explicit discontinuity is enforced upon the skateboard's kick-off into the air using a guard function that reverses angular velocity.

Generated optimal trajectories were limited below a maximum jump height due to constraints on control input acceleration. In future work, system parameters, such as rider mass $m_r$ and skateboard length $L$, could be varied to test if this results in any respective change in maximum jump height. Furthermore, inputs can be more realistically constrained by changing the constant control acceleration limit to more closely follow the specified torque-speed curve of a chosen motor. Other more realistic assumptions could be made to test the real-world accuracy of the abstracted theoretical system. For example, wheels could be modeled with non-zero mass, allowing frictional forces to be included in system dynamics.


\begin{thebibliography}{00}
\bibitem{b1} A. Fried-Cassorla, "The ultimate skateboard book, Running Press," 1988.
\bibitem{b2} "Jumping: the ollie." [Online]. Available:
https://www.exp1oratorium.edu/skateboarding/trick02.html
\bibitem{b3} E.C. Frederick, J.J. Determan, S.N. Whittlesey, and J. Hamill, "Biomechanics of skateboarding: kinetics of the Ollie," J Appl Biomech., pp. 33-40, February 2006.
\bibitem{b4} B. Varszegi, D. Takacs, G. Stepan, and J.S. Hogan, "Stabilizing skateboard speed-wobble with reflex delay," J. R. Soc. Interface, 2016.
\bibitem{b5} M.S. Walsh, C.C. Creekmur, and J. Wojcik, “FORCE TIME MEASURES OF BEGINNING AND SKILLED SKATEBOARDERS PERFORMING AN OLLIE,” 2007.
\bibitem{b6} M. Posa, C. Cantu, and R. Tedrake, "A direct method for trajectory optimization of rigid bodies through contact," The International Journal
of Robotics Research, vol. 33, no. 1, pp. 69-81, 2014. 
\bibitem{b7} R. Tedrake, "Underactuated robotics: Algorithms for walking, running, swimming, flying, and
manipulation (Course Notes for MIT 6.832)", [Online]. Available: http://underactuated.mit.
edu/.
\bibitem{b8} J. Hogg, and J. Scott, "On the effects of scaling on the performance of Ipopt," 2013.
\end{thebibliography}
\end{document}